%% file: main.tex
\documentclass[sigconf,10pt]{acmart}

\usepackage{booktabs} 
\usepackage{xspace}
\usepackage{balance}

\usepackage{macro}

\newcommand{\revised}[1]{#1}

\usepackage{enumitem}

\setlist[itemize,enumerate]{leftmargin=10pt}

\copyrightyear{2019} 
\acmYear{2019} 
\setcopyright{rightsretained} 
\acmConference[SIGMOD '19]{2019 International Conference on Management of Data}{June 30-July 5, 2019}{Amsterdam, Netherlands}
\acmBooktitle{2019 International Conference on Management of Data (SIGMOD '19), June 30-July 5, 2019, Amsterdam, Netherlands}\acmDOI{10.1145/3299869.3314036}
\acmISBN{978-1-4503-5643-5/19/06}
\begin{document}
\title{\drybell: A Case Study in Deploying \\ Weak Supervision at Industrial Scale}

\author{\vspace{-.5em}Stephen H. Bach$^\dagger$\;\;\;\;
Daniel Rodriguez$^\ddagger$\;\;\;\;
Yintao Liu$^\ddagger$\;\;\;\;
Chong Luo$^\ddagger$ \\ \vspace{.25em}
Haidong Shao$^\ddagger$\;\;\;\;
Cassandra Xia$^\ddagger$\;\;\;\;
Souvik Sen$^\ddagger$\;\;\;\;
Alex Ratner$^\mathsection$ \\ \vspace{.25em}
Braden Hancock$^\mathsection$\;\;
Houman Alborzi$^\ddagger$\;\;
Rahul Kuchhal$^\ddagger$\;\;
Chris R{\'e}$^\mathsection$\;\;
Rob Malkin$^\ddagger$\\}

\affiliation{\vspace{1em} $^\dagger$Brown University \;\;\;\;\;\; $^\ddagger$Google \;\;\;\;\;\; $^\mathsection$Stanford University \\ \vspace{.5em}}

\renewcommand{\shortauthors}{S. H. Bach et al.}
\renewcommand{\shorttitle}{Snorkel DryBell}

\input{01_abstract}


\maketitle

\input{10_intro}

\input{20_background}

\input{30_weak-supervision}

\input{40_cross-modal}

\input{50_system}

\input{60_experiments}

\input{70_discussion}

\input{80_related}

\input{90_conclusions}

\begin{acks}
The authors would like to thank Vikaram Gupta, Shiv Venkataraman, and Sugato Basu for their support and help preparing the manuscript.
A.R. gratefully acknowledges the support of the Stanford Bio-X Fellowship. A.R., B.H., and C.R. gratefully acknowledge the support of DARPA under Nos. FA87501720095 (D3M) and FA86501827865 (SDH), NIH under No. N000141712266 (Mobilize), NSF under Nos. CCF1763315 (Beyond Sparsity) and CCF1563078 (Volume to Velocity), ONR under No. N000141712266 (Unifying Weak Supervision), the Moore Foundation, NXP, Xilinx, LETI-CEA, Intel, Google, NEC, Toshiba, TSMC, ARM, Hitachi, BASF, Accenture, Ericsson, Qualcomm, Analog Devices, the Okawa Foundation, and American Family Insurance, and members of the Stanford DAWN project: Intel, Microsoft, Teradata, Facebook, Google, Ant Financial, NEC, SAP, and VMWare. The U.S. Government is authorized to reproduce and distribute reprints for Governmental purposes notwithstanding any copyright notation thereon. Any opinions, findings, and conclusions or recommendations expressed in this material are those of the authors and do not necessarily reflect the views, policies, or endorsements, either expressed or implied, of DARPA, NIH, ONR, or the U.S. Government.
\end{acks}

\bibliographystyle{ACM-Reference-Format}
\balance
\bibliography{bach-sigmod-industrial19}

\end{document}

%% file: 01_abstract.tex
\begin{abstract}
Labeling training data is one of the most costly bottlenecks in developing machine learning-based applications.
\revised{
We present a first-of-its-kind study showing how existing knowledge resources from across an organization can be used as weak supervision in order to bring development time and cost down by an order of magnitude, and introduce \drybell, a new weak supervision management system for this setting.
\drybell builds on the Snorkel framework, extending it in three critical aspects: flexible, template-based ingestion of diverse organizational knowledge, cross-feature production serving, and scalable, sampling-free execution.
On three classification tasks at \webco, we find that \drybell\ creates classifiers of comparable quality to ones trained with tens of thousands of hand-labeled examples, converts non-servable organizational resources to servable models for an average 52\% performance improvement, and executes over millions of data points in tens of minutes.
}
\end{abstract}

%% file: 10_intro.tex
\section{Introduction}
\label{sec:intro}

One of the most significant bottlenecks in developing machine learning applications is the need for hand-labeled training data sets.
In industrial and other organizational deployments, the cost of labeling training sets has quickly become a significant capital expense: collecting labels at scale requires carefully developing labeling instructions that cover a wide range of edge cases; training subject matter experts to carry out those instructions; waiting sometimes months or longer for the full results; and dealing with the rapid depreciation of training sets as applications shift and evolve.

As a result, in industry and other domains there has been a major movement towards programmatic or otherwise more efficient but noisier ways of generating training labels, often referred to as \textit{weak supervision}.
Given the increasing commoditization of standard machine learning model architectures, the supervision strategy used is increasingly the key differentiator for end model performance, and recently has been a key element in state-of-the-art results~\cite{cubuk2018autoaugment,mahajan:eccv18}.
Many prior weak supervision approaches rely on a single source of labels, a small number of carefully chosen, manually combined sources~\cite{mintz2009distant,zhang2016extracting}, or on sources that make uncorrelated errors such as independent crowd workers~\cite{dawid1979maximum,dalvi2013aggregating}.
\revised{
Recent work has explored building end-to-end systems for programmatic supervision, e.g., simple heuristic rules and pattern matching scripts written from scratch by users, which may have diverse accuracies and correlations~\cite{ratner2016data,ratner2017snorkel,bach2017learning}.
}

\begin{figure}[t]
\begin{center}
\centerline{\includegraphics[width=\columnwidth]{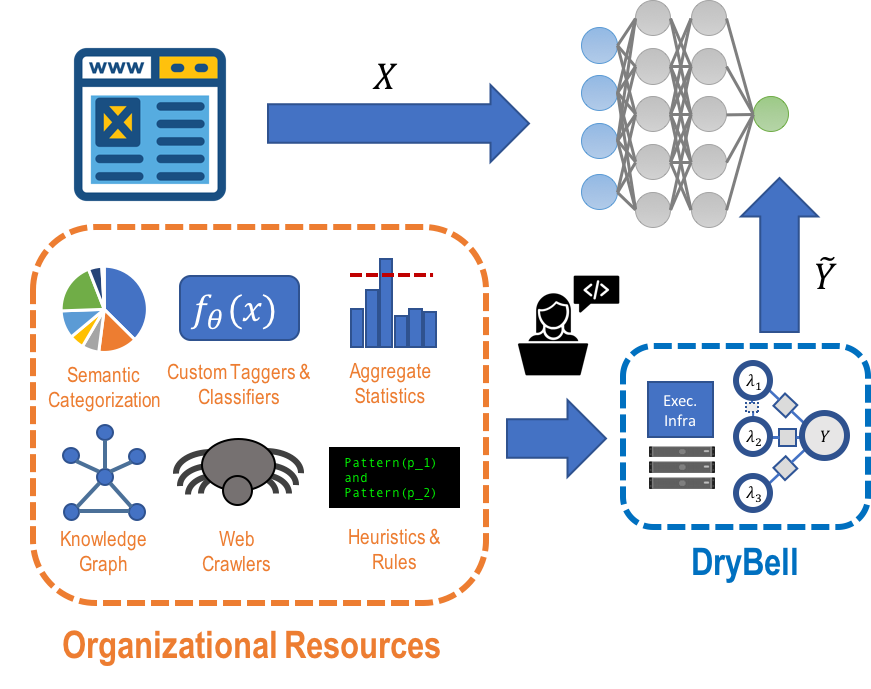}}
\vskip -.1in
\caption{
    \revised{
    Rather than using hand-labeled training data, \drybell uses diverse organizational resources as weak supervision to train content and event classifiers on \webco's platform.
    }
}
\label{fig:example}
\end{center}
\vskip -0.4in
\end{figure}

\revised{
In this work, we present a first-of-its-kind study showing how existing organizational knowledge can be used as weak supervision to have significant impact even in some of the most heavily-engineered, high-value industrial ML applications.
We introduce \drybell, a production-scale \textit{weak supervision management system} which extends the Snorkel framework~\cite{ratner2017snorkel} to handle three novel aspects we find to be critical: flexible template-based ingestion of \textit{organizational knowledge}, \textit{cross-feature} production serving, and scalable, \textit{sampling-free} modeling.
We evaluate \drybell on three content and real-time event classification applications at \webco (Figure~\ref{fig:example}), where we move beyond using simple pattern matchers over text as weak supervision for small-scale, first-cut applications (as in initial work on Snorkel~\cite{ratner2017snorkel}), and demonstrate the value of using existing organizational knowledge resources, measured against baselines with person-decades of development.
}

Based on our experience at Google, we outline three core principles that are central to deploying weak supervision at organizational scale, and highlight how these are implemented in \drybell:
\begin{itemize}
    \item \textbf{\revised{Flexible Ingestion of Organizational Knowledge}}:
    In large organizations, a wide range of resources---such as models, knowledge bases, heuristics, and more---are often available; a weak supervision system should support rapid and flexible integration of as many of these resources as possible for quickly training models to the highest possible quality.
    We highlight the importance of this approach with three case studies involving content and event classification tasks.
    Engineers at \webco\ are responsible for hundreds of separate classifiers, which often rely on hand-labeled training data.
    They must be responsive to everything from shifting business objectives to changes in products; updating these classifiers is often the critical blocker to core product and feature launches.
    We describe how a single engineer can use weak supervision to rapidly develop new classifiers, leading to average, relative quality improvements of 11.5\% (measured in F1 points) over classifiers trained on small $\sim$15K-example development sets, and reaching the quality equivalent of using 80K labels.
    \item \textbf{\revised{Cross-Feature Production Serving}}:
    Organizational\\ knowledge is often present in non-servable form factors, i.e., too slow, expensive, or private to be used in production; instead, a weak supervision system can provide a way to use these to quickly train \textit{servable} models suitable for deployment.
    For example, internal models or heuristics are often defined over features like monthly aggregate statistics, expensive internal models, etc., whereas \drybell can allow users to quickly transfer this knowledge to models defined over \textit{servable} features, e.g., inexpensive, real-time signals.
    We demonstrate how \drybell\ allows users to quickly and flexibly transfer organizational knowledge from non-servable forms to new servable deployment models focused on the classification task of interest.
    We view this as a practical, flexible form of \textit{transfer learning}, and show that incorporating these resources leads to 52\% average gains in performance.
    \item \textbf{\revised{Scalable, Sampling-Free Execution}}:
    A weak supervision system should cleanly decouple \textit{subject matter experts (SMEs)}, who should be able to rapidly and iteratively specify weak supervision, from the details of execution and model training over industrial scale datasets.
    We describe how the architecture of \drybell cleanly decouples the interface by which SMEs across an organization can contribute labeling strategies, and the system for executing these at massive scale while supporting rapid human-in-the-loop iteration---for example, implementing weak supervision over 6M+ data points with sub-30min. execution time---\revised{including a new TensorFlow compute graph-based generative modeling approach that avoids expensive sampling, and a MapReduce-based pipeline}.
\end{itemize}

We achieve these principles in \drybell by adopting the three main stages of the Snorkel pipeline: first, users write \textit{labeling functions}, which are simply black-box functions that take in unlabeled data points and output a label or abstain, and can be used to express a wide variety of weak supervision strategies; next, a generative modeling approach is used to estimate the accuracies of the different labeling functions based on their observed agreements and disagreements; and finally, these accuracies are used to re-weight and combine the labels output by the labeling functions, producing \textit{probabilistic} (confidence-weighted) training labels.

We start in Section~\ref{sec:background} with a brief description of existing work on weak supervision, and of the approach taken by Snorkel, the framework that \drybell extends.
In Section~\ref{sec:weak-supervision}, we present three case studies of content and event classification applications at \webco, where we survey the categories of weak supervision strategies that can be employed within \drybell.
We discuss these case studies at a high level due to the proprietary nature of the applications.
In Section~\ref{sec:cross-modal}, we highlight a particularly critical \revised{\textit{cross-feature}} form of \revised{production model serving} supported in \drybell, in which \textit{non-servable} supervision resources that are expensive to run, private, or otherwise not servable in production are used to train \textit{servable} deployment models.
In Section~\ref{sec:system}, we then present the architecture of \drybell\revised{, emphasizing a new sampling-free generative modeling approach, and a MapReduce-based pipeline and template library}.
In Section~\ref{sec:experiments}, we describe experimental results \revised{on benchmark data sets using Google data representative of production tasks.
We show that \drybell can replace hand-labeling tens of thousands of training examples.}
Finally, we conclude with lessons learned on how weakly supervised machine learning can be integrated into the development processes of production machine learning applications\revised{, and discuss how these lessons can be applied at many different kinds of organizations.}

%% file: 20_background.tex
\section{Background}
\label{sec:background}

In recent years, modern machine learning models have achieved new state-of-the-art accuracies on a range of traditionally challenging tasks.
However, these models generally require massive hand-labeled training sets~\cite{DBLP:journals/corr/SunSSG17}.
In response, many machine learning developers have increasingly turned to \textit{weaker} methods of supervision, in which a larger volume of cheaper, noisier labels is used~\cite{dawid1979maximum,dalvi2013aggregating,mintz2009distant,zhang2016extracting,mnih2012learning,bootkrajang2012label,ratner2017snorkel}.

\revised{
We build on top of Snorkel, a recently proposed framework for weakly supervised machine learning~\cite{ratner2017snorkel}, which allows users to generically specify multiple sources of programmatic weak supervision---such as rules and pattern matchers over text---that can vary in accuracy, coverage, and that may be arbitrarily correlated.}
The Snorkel pipeline follows three main stages, which we also adopt in \drybell: first, users write \textit{labeling functions}, which are simply black-box functions that take in unlabeled data points and output a label or abstain; next, a \textit{generative model} is used to estimate the accuracies of the different labeling functions, and then to re-weight and combine their labels to produce a set of \textit{probabilistic} training labels\revised{, effectively solving a novel data cleaning and integration problem}; and finally, these labels are use to train an arbitrary end \textit{discriminative model}, which is used as the final classifier in production.

This setup can be formalized as follows.
Let $X = (X_1,\dots, X_m)$ be a collection of unlabeled data points, $X_i \in \mathcal{X}$, with associated \emph{unobserved} labels $Y = (Y_1, \dots, Y_m)$.
For simplicity, we focus on binary classification, $Y_i \in \{-1, 1\}$, however \drybell can handle arbitrary categorical targets as well, e.g. $Y_i \in \{1, \ldots, k\}$.

\revised{In our weak supervision setting, we do not have access to these ground-truth labels $Y_i$, and our goal is to estimate them to use as training labels.}
Instead, we have access to $n$ labeling functions $\lambda = (\lambda_1, \dots, \lambda_n)$, where $\lambda_j : \mathcal{X} \rightarrow \{-1, 0, 1\}$, \revised{with $0$ corresponding to an \textit{abstain} vote.}
We use a \textit{generative model} wherein we model each labeling function as abstaining or not with some probability, and labeling a data point correctly with some probability.
Let $\Lambda$ be the matrix of labels output by the $n$ labeling functions over the $m$ unlabeled data points, such that $\Lambda_{i,j} = \lambda_j(X_i)$.
We then estimate the parameters $w$ of this generative labeling model $P_w(\Lambda, Y)$ by maximizing the log marginal likelihood of the observed labels $\Lambda$:
\[
\hat{w} = \argmax_{w} \; \; \log \sum_{Y \in \{-1,1\}^m} P_w(\Lambda, Y) \; .
\]
Note that we are marginalizing out $Y$, i.e. we are not using any ground truth training labels in our learning procedure; instead, we are learning solely from the information about agreements and disagreements between the labeling functions, as contained in the observed label matrix $\Lambda$.
We discuss the choice of the structure of $P_w(\Lambda, Y)$ and the unsupervised approach to estimating $\hat{w}$ further in Section~\ref{sec:gen_model}.

Given the estimated generative model, we use its predicted label distributions, $\tilde{Y}_i = P_{\hat{w}}(Y_i | \Lambda)$, as \textit{probabilistic} training labels for the end \textit{discriminative classifier} that we aim to train.
We train this discriminative classifier $h_\theta$ on our weakly labeled training set, $(X, \tilde{Y})$, by minimizing a \textit{noise-aware} variant of a standard loss function, $l$, i.e. we minimize the expected loss with respect to $\tilde{Y}$:
\[
\hat{\theta} = \argmin_{\theta} \; \; \sum_{i=1}^m \mathbb{E}_{y \sim \tilde{Y}_i}\left[ l(h_\theta(X_i), y) \right]
\]
A formal analysis shows that as the number of unlabeled data, i.e. $m$, is increased, the generalization error of the discriminative classifier should decrease at the same asymptotic rate as it would if supervised with traditional hand-labeled data~\cite{ratner2016data}.
More generally, we expect the discriminative classifier to provide performance gains over the generative model (i.e. the reweighted combination of the labeling function outputs) that it is trained on, both by applying to data types that the labeling functions cannot be applied to, e.g. servable versus non-servable features (see Section~\ref{sec:cross-modal}), and by learning to generalize beyond them.
For example, the discriminative classifier can learn to put weight on more subtle or synonymous features that the labeling functions (and thus, the generative model) do not cover.
For empirical evidence of this generalization, and further details of the actual discriminative models used, see Section~\ref{sec:experiments}.

\revised{
In building \drybell, we sought to extend Snorkel to study weak supervision in the context of an organizational-scale deployment, focusing on three key aspects.
First, while Snorkel was designed to handle ``de novo'' weak supervision applications built with a handful of simple pattern-matching rules, written from scratch by domain experts, we design a template-based interface for ingesting existing organizational knowledge resources like internal models, taggers, legacy code, and more (Section~\ref{sec:weak-supervision}).
Second, we implement support for cross-feature production serving, where weak supervision defined over features not \textit{servable} for application deployment---such as aggregate statistics or expensive results of model inference---can be transferred to deployable models defined over \textit{servable} features (Section~\ref{sec:cross-modal}).
Finally, our design of the \drybell\ architecture focuses on handling massive scale (e.g. 6.5M data points in one application), and thus we focus on speeding up both labeling function execution and generative model training, in particular using a new sampling-free modeling approach (Section~\ref{sec:system}).
}

%% file: 30_weak-supervision.tex
\section{Case Studies: Weak Supervision for Rapid Development}
\label{sec:weak-supervision}

\begin{figure}[t]
\begin{center}
\centerline{\includegraphics[width=\columnwidth]{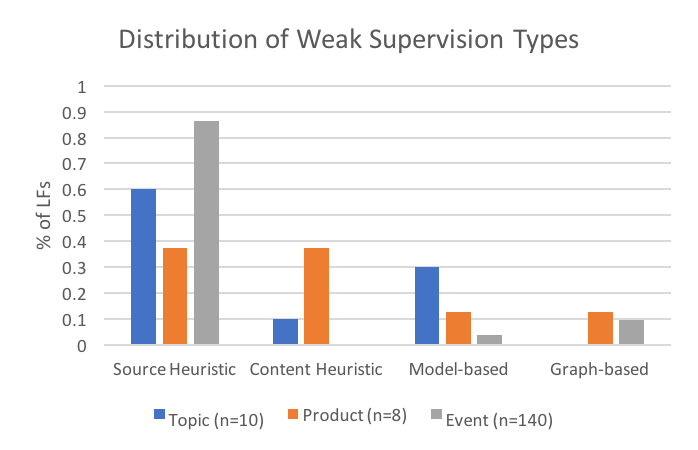}}
\caption{
    We plot the distribution of high-level categories of weak supervision types, counted by number of \textit{labeling functions (LFs)}, for the three applications.
}
\label{fig:ws-types}
\end{center}
\vspace{-1em}
\end{figure}

We start by exploring three case studies of weak supervision applied to classification problems at \webco: two on content classification related to topics and commercial product categories, and one for classifying \appthree\ across several serving platforms.
In this section, we focus on highlighting the diversity of weak supervision signals from across the organization that developers were able to express as \textit{labeling functions (LFs)} in \drybell.
We broadly categorize the weak supervision sources into several coarse-grained buckets, representing different types of organizational knowledge and resources (Figure~\ref{fig:ws-types}):
\begin{itemize}
    \item \textbf{Source Heuristics:} Labeling functions expressing heuristics (pattern or otherwise) about the \textit{source} of the content or event, or aggregate statistics of this.
    \item \textbf{Content Heuristics:} Labeling functions expressing heuristics about the content or event.
    \item \textbf{Model-Based:} Labeling functions that use the predictions of internal models which were developed for some related or component problem.
    Examples include topic models and named entity recognizers applied to content.
    \item \textbf{Graph-Based:} Labeling functions that use a knowledge or entity graph to derive labels.
\end{itemize}
We now describe the applications, giving examples of the above weak supervision source types used in each.

\subsection{Topic Classification}
In the first task, an engineering team for a \webco product needed to develop a new classifier to detect a topic of interest in its content.
The team oversees well over 100 such classifiers, each with its own set of training data, so there is strong motivation for finding faster and more agile ways to develop or modify these models.
Currently, however, the default procedure for developing a new classifier such as this one requires substantial manual data labeling effort.

In our study, we instead used \drybell\ to weakly supervise 684,000 unlabeled data points, selected by a coarse-grained initial keyword-filtering step.
A developer then spent a short time writing ten labeling functions that both expressed basic heuristics, and pulled on organizational resources such as existing models at \webco.
Specific examples of labeling functions included:
\begin{itemize}
    \item \textit{URL-based:} Heuristics based on the linked URL;
    \item \textit{NER tagger-based:} Heuristics over entities tagged within the content, using custom named entity recognition (NER) models maintained internally at \webco;
    \item \textit{Topic model-based:} Heuristics based on a topic model maintained internally at \webco. This topic model output semantic categorizations far too coarse-grained for the targeted task at hand, but which nonetheless could be used as effective negative labeling heuristics.
\end{itemize}

These weak supervision strategies pulled on diverse types of signal from across \webco's organization, but were simple to write within the \drybell\ framework.
With these strategies, we matched the performance of 80K hand-labeled training labels, and get within 4.6 F1 points of a model trained on 175K hand-labeled training data points (see Section~\ref{sec:experiments}).

\subsection{Product Classification}
In a second case study with the same engineering team at \webco, a strategic decision necessitated a modification of an existing classifier for detecting content references to products in a category of interest.
The category of interest was expanded to include many types of accessories and parts---meaning that all previously negative class labels (i.e., ``not in the category of interest'') needed to be relabeled, or else discarded.
In fact, our post-hoc experiments revealed that even using the previously positive labels resulted in a slight reduction in end model F1 score, highlighting the near-instantaneous depreciation of a significant labeling investment given a change in strategy.

Instead, in a similar process to the content classification scenario described above, one developer was able to write eight labeling functions, leveraging diverse weak supervision resources from across the organization.
These labeling functions included:

\begin{itemize}
    \item \textit{Keyword-based:} Keywords in the content indicated either products and accessories in the category of interest, or other accessories not of interest;
    \item \textit{Knowledge Graph-based:} In order to increase coverage across the many languages for which this classifier is used, we queried \webco's Knowledge Graph for translations of keywords in ten languages;
    \item \textit{Model-based:} We again used the semantic topic model to identify content obviously unrelated to the category of products of interest.
\end{itemize}

A classifier trained with these labeling functions matched the performance of 12K hand-labeled training examples, and got within 5.1 F1 points of classifier model trained on 50K hand-labeled training examples (see Section~\ref{sec:experiments}).

\subsection{Real-Time Event Classification}
Finally, we applied \drybell\ to a \appthree\ classification problem over two of \webco's platforms.
In this setting, a common approach is to classify events based on offline (or \textit{non-servable}) features such as aggregate statistics and relationship graphs.
However, this approach induces latency between when an event occurs and when it is identified.
An alternative approach is to use a machine learning model to classify events directly from real-time, \textit{event-level} features.
However, getting hand-labeled training data in this setting is challenging due to the shifting environment, as well as the cost of trained expert annotators.
Instead, we used \drybell\ to train models over the event-level features using weak supervision sources (n=140) defined over the non-servable features, spanning three broad categories:

\begin{itemize}
    \item \textit{Model-based:} Several smaller models that had previously been developed over various feature sets were also used as weak labelers in this setting.
    \item \textit{Graph-based:} A set of models over graphs of entity and destination relationships provided higher recall but generally lower-precision signals than the heuristic classifiers.
    \item \textit{Other heuristics:} A large set of existing heuristic classifiers that had previously been developed.
\end{itemize}

These sources were combined in \drybell\ and used to train a deep neural network over real-time event-level features.
Compared to the same network trained on an unweighted combination of the labeling functions, \drybell\
identifies 58\% more events of interest, with a quality improvement of 4.5\% according to an internal metric.

One of the aspects that we found critical in this setting was the ability of \drybell\ to estimate the accuracies of different labeling functions.
Given the large number of weak supervision sources in play, determining the quality or utility of each source, and tuning their combinations accordingly, would have itself been an onerous engineering task.
Using \drybell, these weak supervision signals could simply all be integrated as labeling functions, and the resulting estimated accuracies were found to be independently useful for identifying previously unknown low-quality sources (which were then either fixed or removed).

%% file: 40_cross-modal.tex
\section{\revised{Cross-Feature Model Serving}}
\label{sec:cross-modal}

\begin{figure}[t]
\begin{center}
\centerline{\includegraphics[width=\columnwidth]{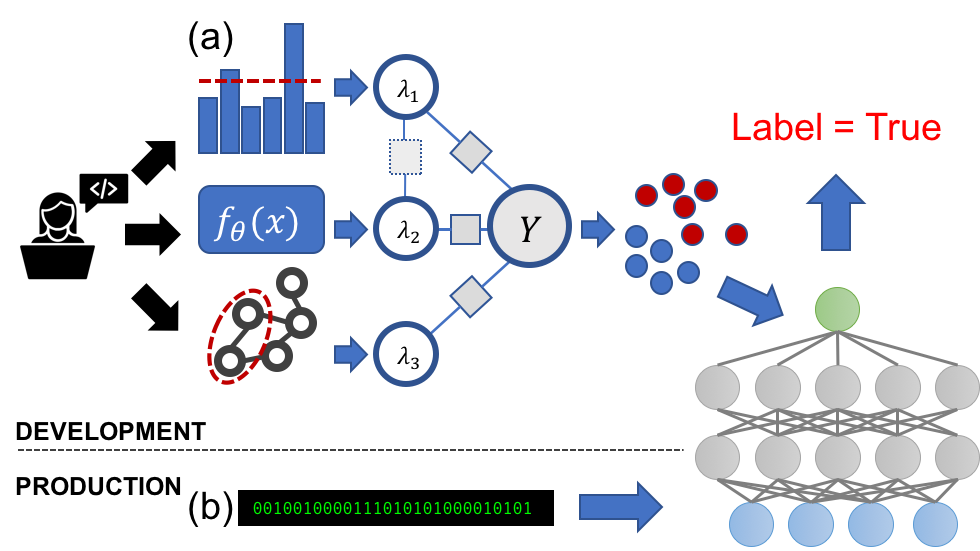}}
\caption{
\revised{
In \drybell, developers can use \textit{non-servable} development features for weak supervision, to train classifiers that operate over separate \textit{servable} features in production.
}
}
\label{fig:x-modal-example}
\end{center}
\vspace{-1em}
\end{figure}

One significant advantage of a weakly supervised approach, as implemented in \drybell, is the ability to easily and flexibly transfer knowledge contained in \textit{non-servable} feature sets that are too slow, expensive, or private to use in production, to \textit{servable} feature sets such as real-time event-level signals or cheap edge-computable features, as in the real-time event application (Figure~\ref{fig:x-modal-example}).
Formally, this goal of transferring knowledge from a model defined over one feature set to a new model trained over another feature set can be viewed as a type of transfer learning~\cite{pan2010survey}, or as similar to a transductive form of model distillation~\cite{hinton2015distilling}.
However, most commonly used transfer learning techniques today apply to models with similar or identical architectures defined over the same basic feature set.
Instead, with \drybell\ we can quickly use models over one set of features---for example, aggregate statistics, results of expensive crawlers, internal models or graphs---and use these to supervise a new model over external, cheap, or otherwise \textit{servable} features.

In the applications we survey at \webco, this is an essential element.
In the \appthree case study, as outlined in the preceding section, none of the weak supervision sources are directly applicable to the event-level, real-time, servable features of interest; instead, with \drybell\ we use them to train a new model that is defined over these servable features.
For the two content applications, while some of the labeling functions could be applied at test time over servable features, others---specifically, those comprising internal models that are expensive to run, or features obtained with high-latency such as the result of web crawlers---are effectively non-servable.
By incorporating the signal from these non-servable sources in \drybell, we get average gains of 52\% in final F1 score performance according to an ablation.
We find that this ability to bridge the gap between non-servable organizational resources and servable models is one of the major advantages of a weak supervision approach like the one implemented in \drybell.

%% file: 50_system.tex
\section{System Architecture}
\label{sec:system}

\begin{figure*}[t]
\begin{center}
\centerline{\includegraphics[width=2\columnwidth]{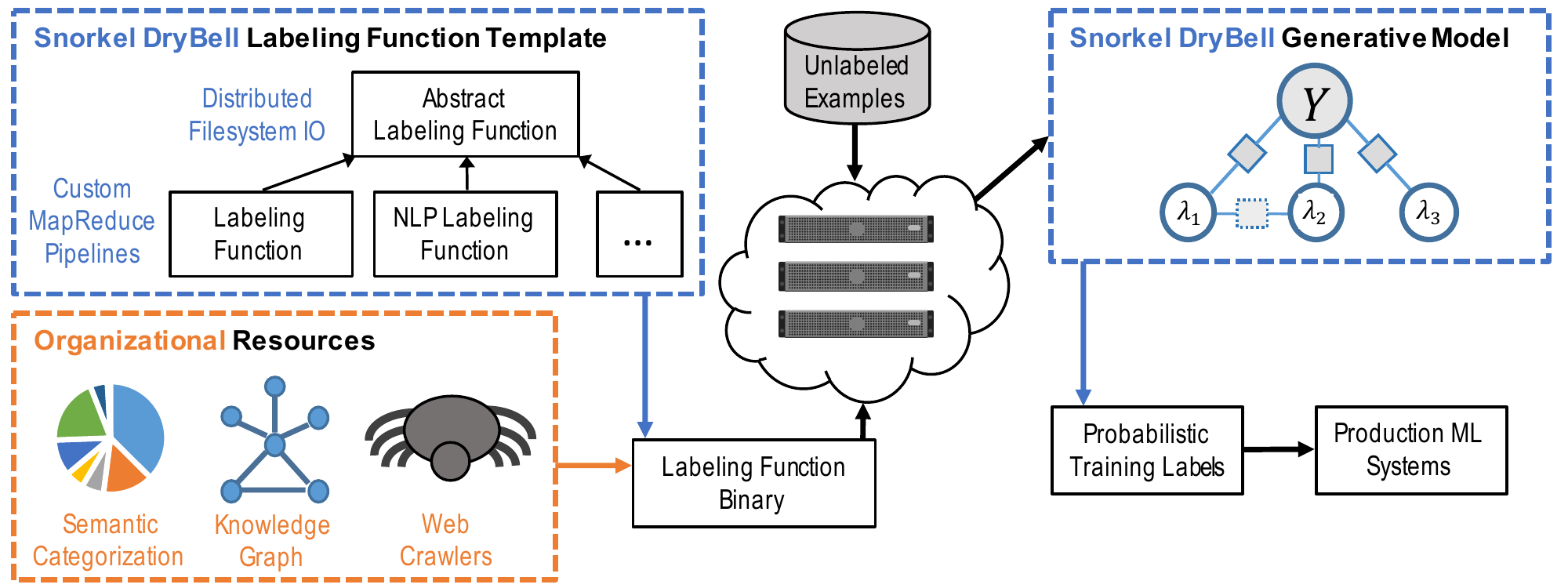}}
\caption{An overview of the \drybell\ system.
(1) \drybell\ provides a library of templated C++ classes, each of which defines a MapReduce pipeline for executing a labeling function with the necessary services, such as natural language processing (NLP).
(2) Engineers write methods for the MapReduce pipeline to determine a vote for each example's label, using \webco\ resources.
(3) \drybell\ executes the labeling function binary on \webco's distributed compute environment.
(4) \drybell\ loads the labeling functions' output into its generative model, which combines them into probabilistic training labels for use by production systems.}
\label{fig:system}
\end{center}
\vspace{-.5em}
\end{figure*}

Deploying the Snorkel framework proposed by Ratner et. al.~\citep{ratner2017snorkel} required redesigning its implementation for an industrial, distributed computing environment, where the scale of examples (millions) is at least an order of magnitude larger than any reported data set for which Snorkel has been used.
This required decoupling and redesigning the labeling function execution and generative modeling components of the pipeline around a template library and distributed compute environment, which we detail next.

\subsection{Labeling Function Template Library}
\label{sec:lf_library}

We implement support for user-defined labeling functions as a library of templated C++ classes.
Our goal is to abstract away the repeated development of code for reading and writing to \webco's distributed filesystem, and for executing MapReduce pipelines.
We achieve this by implementing an \texttt{AbstractLabelingFunction} class that handles all input and output to \webco's distributed filesystem.
Each subclass defines a MapReduce pipeline, with class template slots for functions to be executed within the pipeline.
We initially developed two labeling function pipelines.

The first pipeline is a default pipeline that does not launch any additional services; it simply executes a user-defined function written in C++ (\texttt{LabelingFunction}).
This class meets the needs of many use cases, such as content heuristics, model-based heuristics for models that are executed offline as part of data collection such as semantic categorization, and graph-based heuristics that can query a knowledge graph offline (e.g., categories of products in top-ten languages).

The second pipeline integrates with \webco's general-purpose natural language processing (NLP) models \\ (\texttt{NLPLabelingFunction}).
Such integration is necessary because these NLP models are too computationally expensive to run for all content submitted to \webco.
\drybell\ therefore needs to enable labeling-function writers to execute additional models in a manner that scales to the millions of examples to be labeled.
To achieve this goal, \drybell\ uses \webco's MapReduce framework to launch a model server on each compute node.
Other model servers besides the NLP models can be supported by creating new subclasses of \texttt{AbstractLabelingFunction}.

Engineers using this library need to write only simple main files that define the function(s) that computes the labeling function's vote for an individual example.
These functions capture the engineer's knowledge about how to use existing resources at \webco\ as heuristics for weak supervision.
As an example that is analogous to a labeling function in our content classification application, suppose our goal is to identify content related to celebrities. 
A developer can implement a heuristic that uses a named-entity recognition model for this task as an instance of \texttt{NLPLabelingFunction}.
The labeling function labels any content that \emph{does not} contain a person as \emph{not} related to celebrities.
The first template argument is a pointer to a function that takes an example object as input and selects the text to be provided to the NLP model server.
The second template argument is a pointer to a function that takes the same example object and the output of the NLP models as its inputs, and checks whether the named-entity recognition model found any proper names of people.
We illustrate this example in code:

\begin{cc}
    string GetText(const Example& x) {
        return StrCat(x.title, " ", x.body);
    }
    
    LFVote GetValue(const Example& x,
                    const NLPResult& nlp) {
        if (nlp.entities.people.size() == 0) {
            return NEGATIVE;
        }
        else { return ABSTAIN; }
    }
    
    int main(int argc, char *argv[]) {
        Init(argc, argv);
        NLPLabelingFunction<&GetText, &GetValue> lf;
        lf.Run();
    }
\end{cc}
This short bit of code captures a logical relationship between an existing model and the target task, speeding development.

\revised{\subsection{Sampling-Free Generative Model}}
\label{sec:gen_model}

The critical task in \drybell\ is to combine the noisy votes of the various labeling functions into estimates of the true labels for training.
\revised{
In \drybell, we use a new sampling-free modeling approach which is far less CPU intensive and far simpler to distribute across compute nodes.
}
We focus on a conditionally independent generative model, which we write as:
\begin{align*}
P_w(\Lambda, Y) = \prod_{i = 1}^m P_w(Y_i) \prod_{j = 1}^n P_w(\lambda_j(X_i) | Y_i) \; ,
\end{align*}
\revised{
Following prior work~\cite{ratner2017snorkel}, we assume each labeling function has an accuracy given that it did not abstain, and a propensity to not abstain, i.e., we share parameters across the conditional distributions.
For simplicity, here we assume that $P_w(Y_i)$ is uniform, but we can also learn this distribution.

The learning objective of the generative model is to minimize the negative marginal log-likelihood of the observed labeling function outputs $-\log P(\Lambda)$.
The open-source Snorkel implementation\footnote{\url{http://snorkel.stanford.edu}} uses a Gibbs sampler to compute the gradient of this likelihood, but sampling is relatively CPU intensive and complicated to distribute across compute nodes.
Instead, we design a new TensorFlow-based~\cite{abadi:osdi16} implementation for sampling-free optimization.
For numeric stability, we represent the model parameters in log space.
Let $\alpha_j$ be the unnormalized log probability that labeling function $j$ is correct given that it did not abstain, and let $\beta_j$ be the unnormalized log probability that it did not abstain.
Then, to define a static compute graph, as required by TensorFlow, we use 0-1 indicator functions for each possible label value and multiply by the corresponding log likelihood:
\[
-\log P(\Lambda) = -\sum_{i=1}^m \log \left(
P(\Lambda_i, Y_i=1) + P(\Lambda_i, Y_i=-1)
\right) \, ,
\]
where
\begin{align*}
\log P(\Lambda_i, Y_i=1) =
\sum_{j=1}^n (
&{\boldsymbol 1}[\lambda_j(X_i) = 1](\alpha_j + \beta_j - Z_j) \\
+ &{\boldsymbol 1}[\lambda_j(X_i) = -1](-\alpha_j + \beta_j - Z_j) \\
- &{\boldsymbol 1}[\lambda_j(X_i) = 0] Z_j ) \, , \\
\\
\log P(\Lambda_i, Y_i=-1) =
\sum_{j=1}^n (
&{\boldsymbol 1}[\lambda_j(X_i) = 1](-\alpha_j + \beta_j - Z_j) \\
+ &{\boldsymbol 1}[\lambda_j(X_i) = -1](\alpha_j + \beta_j - Z_j) \\
- &{\boldsymbol 1}[\lambda_j(X_i) = 0] Z_j )\, , \\
\\
Z_j = \log (&\exp(\alpha_j + \beta_j) + \exp(-\alpha_j + \beta_j) + 1 ) \, .
\end{align*}
}

The result is a fast implementation that can take hundreds of gradient steps per second on a single compute node.
For example, in our product classification application, in which there are ten labeling functions, the optimizer takes an average $>100$ steps per second with a batch size of 64.
With ten labeling functions and a batch size of 64, \revised{a Gibbs sampler} averages $<50$ examples per second, so \drybell\ provides a 2$\times$ speedup.
\revised{Implementing the generative model as a static compute graph in TensorFlow has another advantage over a Gibbs sampler.
It is easy to parallelize across multiple compute nodes using TensorFlow's API.
(Here we report timing using a single process for a fair comparison.)
}

It is also possible to relax the conditional independence assumption by defining model functions in TensorFlow that capture specific, low-tree-width graphical model structures, which we leave for future work.
It is also possible to directly plug-in matrix factorization models of the kind recently used for denoising labeling functions \cite{ratner2019training} as TensorFlow model functions.

\revised{
\subsection{Discriminative Model Serving}
To create discriminative models that are servable in production, we integrated \drybell with TFX~\cite{baylor:kdd17}, Google's platform for end-to-end production-scale machine learning.
The probabilistic training labels estimated by \drybell are passed to TFX, where users can configure a model to train with a noise-aware loss function.
Once trained, we use TFX to automatically stage it for serving.

As described in Section~\ref{sec:cross-modal}, the discriminative model acts on a more compact feature representation than the labeling functions, enabling a cross-feature transfer of knowledge from non-servable resources used in labeling functions to a servable model.
TFX supports both logistic regression and deep neural network models, which can operate on user-specified features that are available in production, or on the ``raw'' content, e.g., an LSTM~\cite{hochreiter:neuralcomputation97} that embeds each token of text in a vector space.
}

\subsection{Comparison with Snorkel \revised{Architecture}}

There are several other key differences between \drybell and Snorkel's existing open-source implementation \revised{beyond} the changes detailed above.
Snorkel is designed to run on a single, shared-memory compute node.
In contrast, \webco, like many large organizations, uses a distributed job scheduling and accounting system for large-scale computing.
It therefore was necessary to integrate \drybell\ with \webco's MapReduce framework.

Further, Snorkel is designed to be accessible to novice programmers with limited machine learning experience.
It uses a Jupyter notebook interface and enforces a strict \emph{context hierarchy} data model for representing training data.
This rigid approach is not appropriate for the wide range of tasks that arise in a large organization.
Snorkel also uses a relational database backend for storing data, which does not easily integrate with \webco's existing data-storage systems.
We therefore developed the more loosely coupled system described above, in which labeling functions are independent executables that use a distributed filesystem to share data.

%% file: 60_experiments.tex
\section{Experiments}
\label{sec:experiments}

To evaluate the performance of \drybell, we created benchmark data sets using \webco\ data representative of the production tasks described in Section~\ref{sec:weak-supervision}.
We first show results on the content classification applications, and use them to illustrate trade-offs between weak supervision and collecting hand-labeled data, as well as the benefits of using non-servable features for weak supervision.
We then show results on the \appthree\ application.
\revised{
Due to the sensitive nature of these applications, we report relative improvement to our baselines for the content classification applications. We are unable to describe the details of internal metrics used to evaluate \appthree, but include a high-level description as corroborating evidence that \drybell is widely applicable.}

\begin{table}[t]
\centering
\caption{Number of unlabeled examples used during training $n$,
number of labeled examples in the development set $n_\text{Dev}$ and test set $n_\text{Test}$, fraction of positive labels in $n_\text{Test}$, and number of labeling functions used for each task, for the content classification applications.}
\label{tab:dataset_stats}
\begin{tabular}{l r r r r r}
\toprule
Task & $n$ & $n_{\text{Dev}}$ & $n_{\text{Test}}$ & \% Pos. & \# LFs \\
\midrule
Topic Classification & 684K & 11K & 11K & 0.86 & 10 \\
\AppTwo & 6.5M & 14K & 13K & 1.48 & 8 \\
\bottomrule
\end{tabular}
\end{table}

\subsection{Topic and \AppTwo}

\begin{table*}[t]
\centering
\caption{
    Evaluation of \drybell on content classification tasks, optimizing for F1 score.
    We report numbers relative to the baseline of training directly on the hand-labeled development set.
    Reported scores are normalized relative to the precision, recall, and F1 scores of these baselines, using a true/false threshold of 0.5 for prediction.
    Lift is reported relative to the baseline F1.
    We compare the generative model of \drybell, i.e., a weighted combination of the labeling functions, and the discriminative logistic regression classifier trained with \drybell.
    Note that the generative model is not servable, i.e., it cannot be used to make predictions in production.
}
\label{tab:benchmarks}
\begin{tabular}{l cccr  c cccr}
\toprule
& \multicolumn{4}{c}{Generative Model Only} & \phantom{a} & \multicolumn{4}{c}{\drybell}\\
Task \hfill Relative: & P & R & {\bf F1} & {\bf Lift} & & P & R & {\bf F1} & {\bf Lift}\\
\midrule
Topic Classification    & 84.4\% & 101.7\% & 93.9\% & \textit{-6.1\%} & & 100.6\% & 132.1\% & 117.5\% & \textit{+17.5\%} \\
\AppTwo   & 103.8\% & 102.0\% & 102.7\% & \textit{+2.7\%} & & 99.2\% & 110.1\% & 105.2\% & \textit{+5.2\%} \\
\bottomrule
\end{tabular}
\end{table*}

To evaluate on the topic and \apptwo\ applications, we used the probabilistic training labels estimated by \drybell\ to train logistic regression discriminative classifiers with servable features similar to those used in production.
We have access to hundreds of thousands to millions of unlabeled examples for these tasks.
We also create a small, hand-labeled \textit{development set} ($n_{\textrm{Dev}}$ in Table~\ref{tab:dataset_stats}) which is used by the developer while formulating labeling functions, for hyperparameter tuning of the end discriminative classifier, and as a supervised learning baseline in our experiments.

We use a logistic regression model in TFX.
We train using the FTLR optimization algorithm~\citep{mcmahan:kdd13}, a variant of stochastic gradient descent that tunes per-coordinate learning rates, with an initial step size of 0.2.
We train for 10K iterations for the topic classification task and 100K iterations for the \apptwo\ task, in order to have a similar training time to production classifiers.
(The topic classification task has an order-of-magnitude more features than the \apptwo\ task.)
All experiments use a batch size of 64.

Table~\ref{tab:benchmarks} shows the results of applying the \drybell\ system on the product and topic classification tasks.
We report all results relative to the baseline approach of training the discriminative classifier directly on the hand-labeled development set.

We also report the predictive accuracy of \drybell's generative model, i.e., using the weighted combination of labeling functions directly to make predictions.
We do so to demonstrate that the discriminative classifier learns to generalize beyond the information contained in the labeling  functions.
Note that the generative model is not actually viable for production tasks, because labeling functions often depend on non-servable features of the data.

The results show that on both tasks, the discriminative classifiers trained on \drybell-produced training data has higher predictive accuracy in F1 score on the test sets than classifiers trained directly on the development set.
The weakly supervised classifiers also have higher predictive accuracy than the corresponding generative models.
This result demonstrates that \drybell\ effectively transfers the knowledge contained in the non-servable resources to classifiers that only depend on servable features.

\subsection{Trade-Off Between Weak Supervision Hand-Labeled Data}

\begin{figure}[t]
\begin{center}
\center{
    \includegraphics[width=\columnwidth]{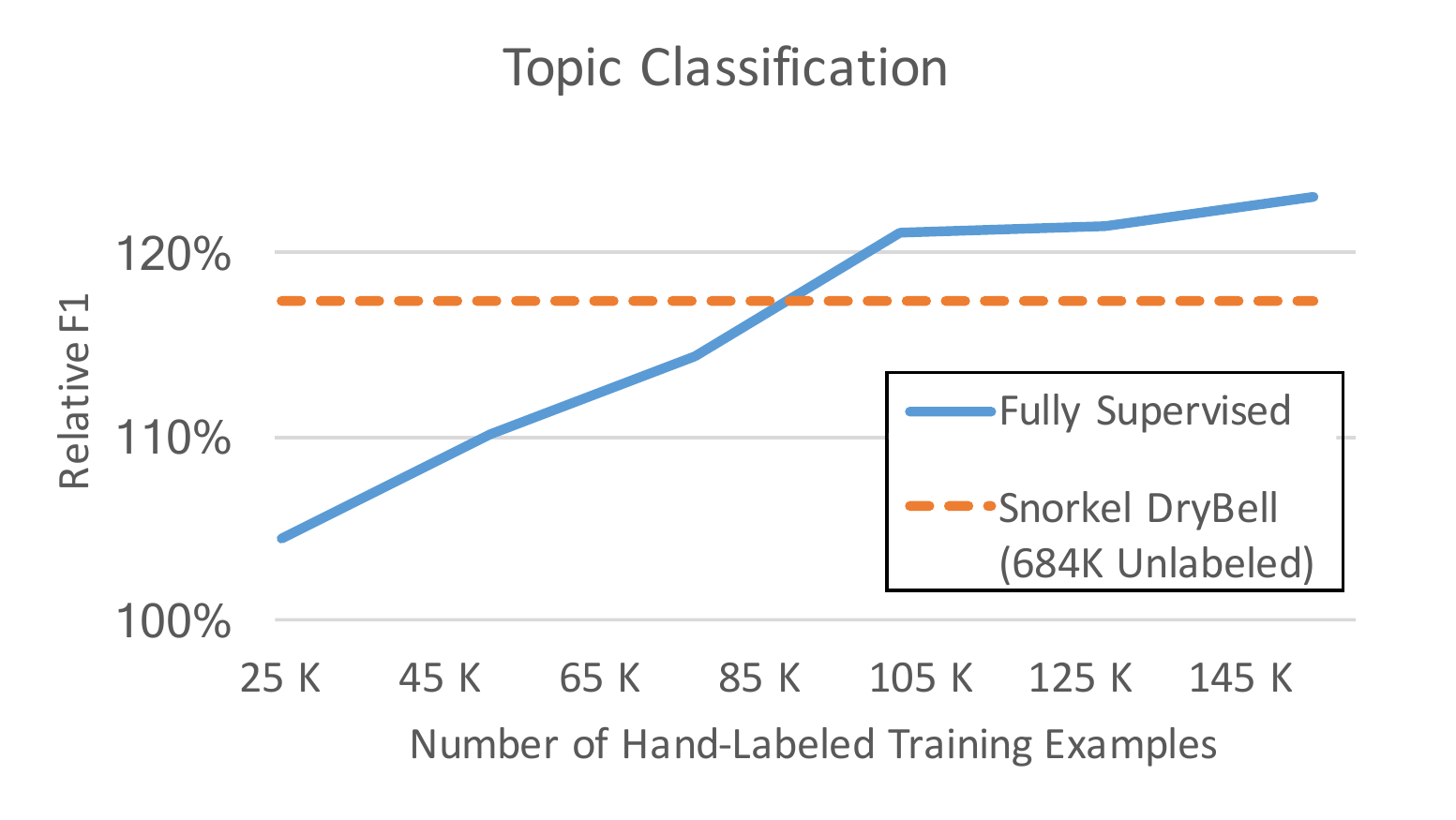} \\
    \vspace{.25in}
    \includegraphics[width=\columnwidth]{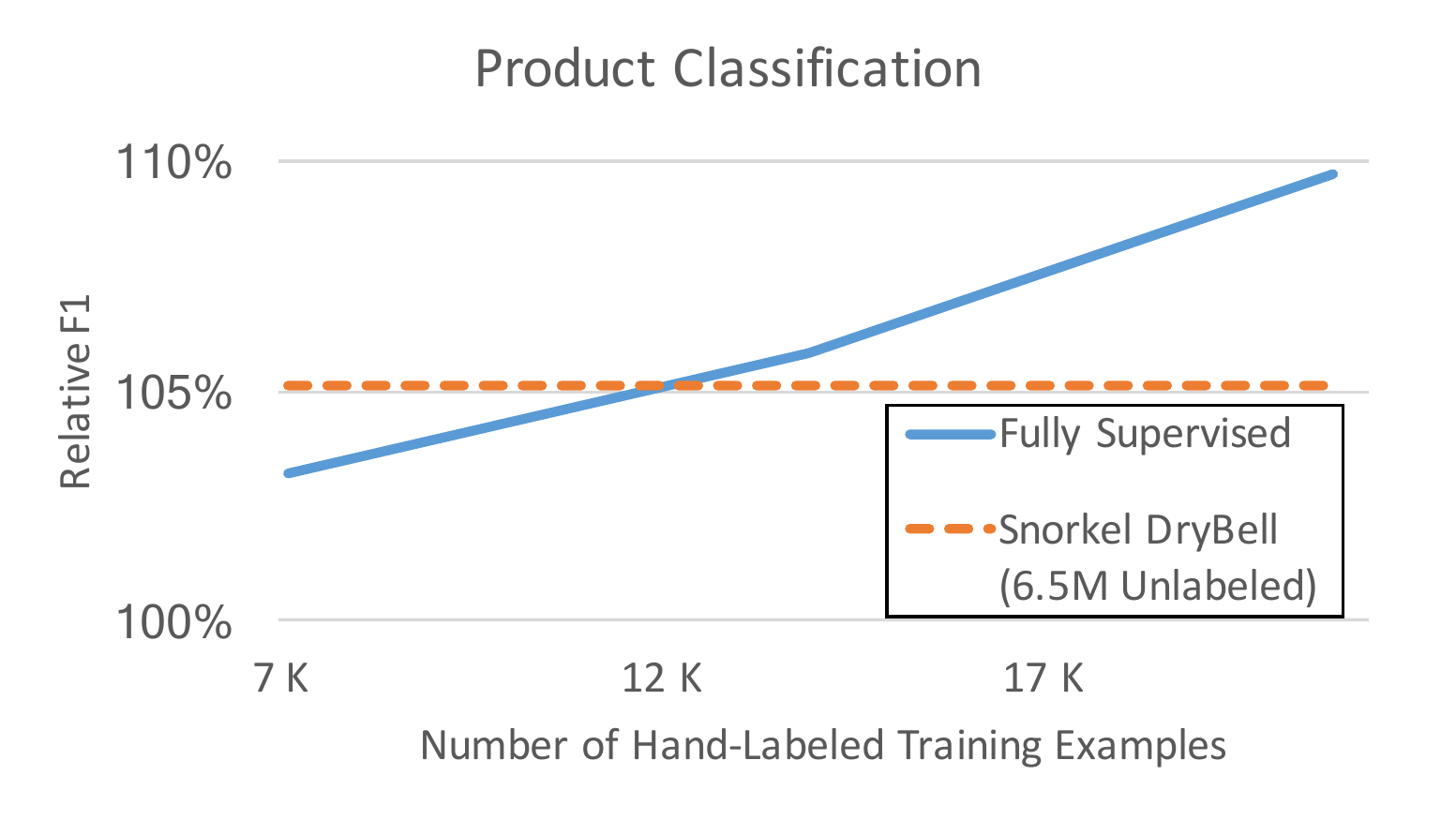}
}
\caption{
   Relative difference in predictive accuracy measured in F1 of supervised classifiers trained on increasing numbers of hand-labeled training examples for the topic and \apptwo\ tasks.
   The dashed line shows the normalized F1 score of the weakly supervised classifier trained on \drybell-inferred labels.
}
\label{fig:trade-off}
\end{center}
\end{figure}

We next investigate the trade-off between using weak supervision and collecting hand-labeled training examples.
We train the discriminative classifier for each content classification task on increasingly large hand-labeled training sets.
Figure~\ref{fig:trade-off} shows the predictive performance in relative F1 score of the the classifiers versus the number of hand-labeled training examples.
On the topic classification task, we find that it takes roughly 80K hand-labeled examples to match the predictive accuracy of the weakly supervised classifier.
On the \apptwo\ task, we find that it takes roughly 12K hand-labeled examples.
This result shows that weak supervision can significantly reduce the need for hand-labeled training data in content classification applications.

\subsection{Ablation Study}

\begin{table}[t]
\centering
\caption{An ablation study of \drybell\ using only labeling functions that depend on servable features (``Servable LFs'') compared with all labeling functions, including non-servable resources.
All scores are normalized to the precision, recall, and F1 of the logistic regression classifier trained directly on the development set. Lift is reported relative to Servable LFs.}
\label{tab:ablation}
\begin{tabular}{l cccr}
\toprule
\hfill Relative: & P & R & {\bf F1} & {\bf Lift} \\
Topic Classification & & & & \\
\midrule
Servable LFs        & 50.9\% & 159.2\% & 86.1\% & \\
+ Non-Servable LFs  & 100.6\% & 132.1\% & 117.5\% & \textit{+36.4\%} \\
\midrule
\AppTwo & & & & \\
\midrule
Servable LFs        & 38.0\% & 119.2\% & 62.5\% & \\
+ Non-Servable LFs  & 99.2\% & 110.1\% & 105.2\% & \textit{+68.2\%} \\
\bottomrule
\end{tabular}
\end{table}

We measured the importance of including non-servable organizational supervision resources by removing all labeling functions that depend on them from the topic and \apptwo\ applications.
The only labeling functions that remained were pattern-based rules.
Table~\ref{tab:benchmarks} shows the results.
We find that incorporating non-servable \webco\ resources in labeling functions leads to a 52\% average performance improvement for the end discriminative classifier.
This result shows that the non-servable resources contain valuable information that are effectively transferred.

\begin{table}[t]
\centering
\caption{An ablation study of \drybell\ using equal weights for all labeling functions to label training data (``Equal Weights'') compared with using the weights estimated by the generative model.
All scores are normalized to the precision, recall, and F1 of the logistic regression classifier trained directly on the development set. Lift is reported relative to Equal Weights.}
\label{tab:ablation2}
\begin{tabular}{l cccr}
\toprule
\hfill Relative: & P & R & {\bf F1} & {\bf Lift} \\
Topic Classification & & & & \\
\midrule
Equal Weights        & 54.1\% & 163.7\% & 109.0\% & \\
+ Generative Model  & 100.6\% & 132.1\% & 117.5\% & \textit{+7.7\%} \\
\midrule
\AppTwo & & & & \\
\midrule
Equal Weights        & 94.3\% & 110.9\% & 103.24\% & \\
+ Generative Model  & 99.2\% & 110.1\% & 105.2\% & \textit{+1.9\%} \\
\bottomrule
\end{tabular}
\end{table}

We also measured the importance of using the generative model to estimate the weights of the labeling function votes by training an identical logistic regression classifier giving equal weight to all the labeling functions for the topic and \apptwo\ applications.
In other words, the probabilistic training labels were an unweighted average of the labeling function votes.
Table~\ref{tab:ablation2} shows the results.
We find that using the generative model to weight labeling functions leads to a 4.8\% average performance improvement for the end discriminative classifier.
This result shows that the generative model is an effective component of the \drybell pipeline.

\subsection{\AppThree}

\begin{figure}[t]
\begin{center}
\centerline{
    \includegraphics[width=0.475\columnwidth]{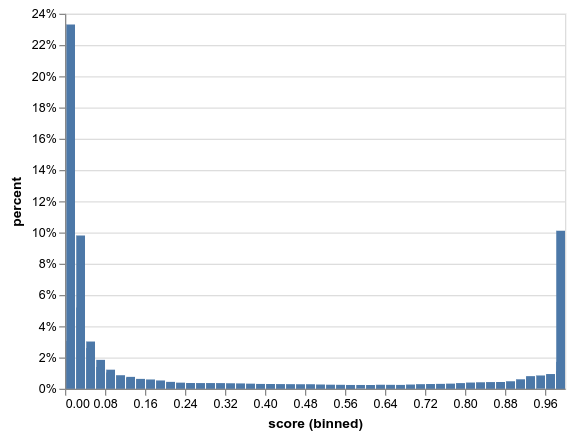}
    \includegraphics[width=0.475\columnwidth]{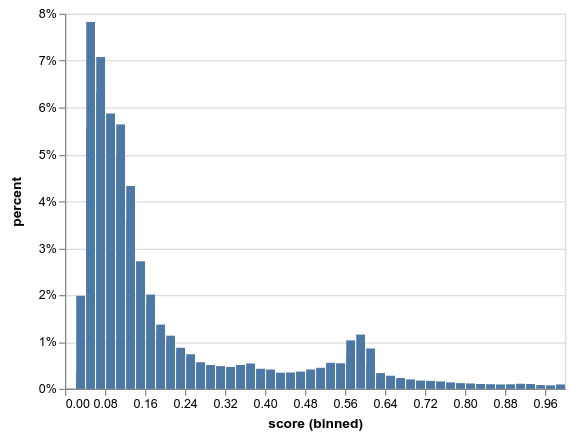}
}
\caption{
    We compare a histogram of the predicted probabilities (``scores'') of an event using a model trained with a baseline Logical-OR approach to combining weak supervision sources (left), and trained using \drybell's output (right).
    We see that the baseline approach results in greatly over-estimating the score of events, whereas the model trained using \drybell\ produces a smoother distribution.
    This results in better performance, and offers more useful output to those monitoring the system.
}
\label{fig:app3-scores}
\end{center}
\vspace{-1em}
\end{figure}

We evaluate the application of \drybell\ to the \appthree\ application as compared to a baseline weak supervision approach of training the same deep neural network architecture on a simpler combination of the same set of labeling functions. Specifically, we compare:
\begin{itemize}
    \item \textit{Logical-OR Weak Supervision:} Here, the weak supervision sources, defined over the non-servable features, are combined using a logical OR. The resulting labels are then used to train a deep neural network (DNN) discriminative classifier over the servable features.
    \item \textit{\drybell}: Here, we use \drybell to combine the weak supervision sources, and then use the resulting probabilistic training labels to train a DNN over the servable features.
 \end{itemize}

We observed that \drybell\ identifies an additional 58\% of events of interest as compared to what the baseline Logical-OR approach captures, and the quality of the events identified by \drybell\ offer a 4.5\% improvement according to an internal metric.

Finally, we note that \drybell\ leads to an end discriminative classifier that produces a more reasonable distribution of \textit{scores}, i.e. predicted probabilities of a certain event label, as compared to the Logical-OR weak supervision baseline (Figure~\ref{fig:app3-scores}).
Whereas the DNN trained using the latter approach ends up predicting labels with nearly absolute confidence, the distribution produced by \drybell\ is far more nuanced and consistent with the expected distribution---resulting not only in better quality, but more interpretable and usable end predictions.

%% file: 70_discussion.tex
\section{Discussion}
\label{sec:discussion}

The experimental results above demonstrate the importance of \drybell's design principles to deploying weakly supervised machine learning in an industrial setting.
First, the ability to incorporate diverse organizational resources was critical to both the content and real-time event applications.
In both cases, \webco\ has a variety of tools from which we constructed weak supervision sources, from existing machine learning classifiers, to structured background knowledge, to previously developed heuristic functions.
These tools are heterogeneous, not just in the information they contain, but how they are maintained and executed within \webco.
Some, like semantic categorization, are maintained by one team and applied generally to incoming content.
Others, like the natural language processing models, are maintained by another team and must be executed as part of the weak supervision development process because they are too expensive to run on all incoming content.
We find that labeling functions are an effective abstraction for encapsulating all these types of heterogeneity.

Second, we find that the mechanism of denoising labeling functions to produce training data and train new classifiers used in \drybell\ effectively transfers knowledge from non-servable resources to servable models.
This is crucial in an industrial environment in which products are composed of many services that are connected via latency agreements.
When engineers have to ensure that classifiers make predictions within allotted times, they have to be very selective about what features to use.
In contrast, writing labeling functions affords developers flexibility because they are executed as part of an offline training process.

Third, we find that the labeling function abstraction is user friendly, in the sense that developers in the organization can write new labeling functions to capture domain knowledge.
\drybell's architecture is designed for high throughput, enabling rapid human-in-the-loop development of labeling functions.
For example, developing each content classification application was possible because of the ability to rapidly iterate on labeling functions.
In contrast, waiting for human annotators to hand-label training data can cause lengthy delays.

We anticipate that this low-latency development of machine learning classifiers will be increasingly important as businesses and other large organizations increasingly depend on machine learning.
This is because machine learning teams are now responsible for implementing business strategies.
For example, if a company like \webco\ decides to add a feature to a product that requires identifying content on a specific topic, the machine learning team currently must respond by curating training examples for this topic.
If the strategy changes, then the training examples must change too.
Weakly supervised machine learning systems like \drybell enable these teams to respond by writing code, rather than pushing high-latency tasks to data annotators.
When launch schedules for products that depend on machine learning are short, time spent curating training data is costly.

This code-as-supervision paradigm also has the potential to meet additional challenges that modern machine learning teams face.
A single team in a large organization is often now responsible for hundreds or more different classifiers.
Each one currently needs its own hand-labeled set of training examples.
We have demonstrated that \drybell\ enables \webco\ to leverage existing resources---including other machine learning classifiers---to create new classifiers.
We anticipate that the problem of managing these large networks of classifiers that share knowledge will be a significant area of future work in the near future.

Finally, we believe weakly supervised machine learning has the potential to affect organizational structures.
\webco\ is beginning to experiment with reorganizing machine learning development around the separation between subject matter expertise and infrastructure enabled by weak supervision.
Dedicated teams could potentially focus on writing labeling functions while others stay focused on serving the resulting classifiers in production.

\revised{
\subsection{Lessons for Other Organizations}
As the code-as-supervision paradigm grows in use, we anticipate that several lessons learned from our work are adoptable beyond Google.
First, resources that can be used for weak supervision are abundant, and our findings demonstrate that labeling functions combined via a generative model are a new way to extract value from them.
In our work, access to broad-purpose natural language processing models saved significant developer time, even though these models were not designed specifically for our tasks.
We observe other organizations already developing broad-purpose models for domains such as language modeling~\cite{grave:neurips17, gupta:arxiv18}, object recognition~\cite{mahajan:eccv18}, and object detection~\cite{zhai:www17}.
In addition, cloud-service providers are making pre-trained, broad-purpose models for language and vision available to consumers \cite{amazon:comprehend, google:cloudai}.
Further, many open-source, broad-purpose models for tasks like named entity recognition~\cite{spacy2}, sentiment analysis~\cite{manning:acl14}, and object detection~\cite{onnx} are freely available and could be incorporated into labeling functions for more specific tasks.
While our study focused on existing resources, our results also indicate that further investment in broad-purpose models and knowledge graphs that provide background knowledge for weak supervision will be increasingly worthwhile.

Second, cross-feature transfer to servable models was critical in our applications, and represents a new perspective on model serving strategies which we believe may be of general applicability.
In this approach, developers use a set of features not servable in production---for example, expensive internal models or private entity network graphs---to create training data for models that are defined over production-servable features.
We learned that having multiple representations of the same data is an effective way to weakly supervise models with service-level agreements.
This technique can potentially benefit the many applications where efficient model serving is needed.

Third, the design choices made in the original Snorkel implementation can be improved for many use cases.
Some changes are applicable to both systems for novice users and experts.
We found that implementing the generative model in an optimization framework with automatic differentiation was faster to develop, easier to distribute, and faster to execute than an MCMC sampling approach.
Other lessons came from the different needs of advanced machine learning engineers versus novice users.
We found that advanced users want maximum flexibility in implementing labeling functions, including being able to launch additional services and call remote procedures during execution.
This approach is in contrast to Snorkel's focus on novice users, in which data to be labeled is represented with a prescribed class hierarchy.
The tradeoff is that while \drybell\ offers fewer higher-level helper functions for labeling function writers, it was easy to apply to a wider range of data, from online content to events.
We anticipate that other implementations of weak supervision for advanced users will want to follow \drybell's approach.
}

%% file: 80_related.tex
\section{Related Work}
\label{sec:related}

Weakly supervised machine learning as implemented in \drybell---using multiple noisy but inexpensive sources of labels as an alternative to hand-labeled training data---is related to other areas of research in machine learning and data systems that also seek to learn and make inferences with limited labeled data.

In machine learning, \emph{semi-supervised learning}~\citep{chapelle:06} combines labeled data with unlabeled data.
It is a broad category of methods that generally seek to use the unlabeled data to discover structure in the data, such as dense clusters or low-dimensional manifolds, that enables better extrapolation from the limited labeled examples.
\textit{Transfer learning}~\cite{pan:tkde10} exploits labeled data available for one or more tasks to reduce the need for labeled data for a new task.
Methods in which a learner labels additional data for itself to train on include \emph{self-training}~\citep{scudder:infotheory65, agrawala:infotheory70}, \emph{co-training}~\citep{blum:colt98}, and \emph{pseudo-labeling}~\citep{lee:workshop13}.
\emph{Zero-shot learning}~\citep{xian:pami18} attempts to learn a sufficiently general mapping between class descriptions and labeled examples that new classes can be identified at test time from just a description, without any additional training examples.
\emph{Active learning}~\citep{settles:12} methods select data points for human annotators to label.
They aim to minimize the amount of labeling needed by interleaving learning and requests for new labels.

Related problems in data systems include \emph{data fusion}~\citep{dong:book15, rekatsinas:sigmod17} and \emph{truth discovery}~\citep{li:kddex15}.
Here the goal is to estimate the accuracy of possibly conflicting records in different data sources and integrate them into the most likely set of correct records.
A similar problem is \emph{data cleaning}~\citep{rahm:dataeng00}, which aims to identify and correct errors in data sets.
Recently, Rekatsinas et. al.~\citep{rekatsinas:vldb17} proposed HoloClean, which uses weakly supervised machine learning to learn to correct errors.
Many methods for these problems, e.g., the latent truth model~\citep{zhao:vldb12}, use generative models similar to the one in \drybell\ in that they represent the unobserved truth as a latent variable.
\drybell's generative model differs in that it models the output of labeling functions executed on input data, and these functions can provide any class label or abstain.

%% file: 90_conclusions.tex
\section{Conclusion}
\label{sec:conclusions}

In this paper we presented the first results from deploying the \drybell\ framework for weakly supervised machine learning in a large-scale, industrial setting.
We find that weak supervision can train classifiers that would otherwise require tens of thousands of hand-labeled examples to obtain, and that \drybell's design enables developers to effectively connect a wide range of organizational resources to new machine learning problems in order to improve predictive accuracy.
These results indicate that weak supervision has the potential to play a significant role in industrial development of machine learning applications in the near future.